# 2nd Place Solution to Google Landmark Retrieval Competition 2020

Min Yang*, Cheng Cui*, Xuetong Xue*, Hui Ren*, Kai Wei*

{yangmin09, cuicheng01, xuexuetong, renhui, weikai}@baidu.com

## Abstract

*This paper presents the 2nd place solution to the Google Landmark Retrieval Competition 2020. We propose a training method of global feature model for landmark retrieval without post-processing, such as local feature and spatial verification. There are two parts in our retrieval method in this competition. This training scheme mainly includes training by increasing margin value of arcmargin loss and increasing image resolution step by step. Models are trained by PaddlePaddle[1] framework and Pytorch[2] framework, and then converted to tensorflow 2.2. Using this method, we got a public score of 0.40176 and a private score of 0.36278 and achieved 2nd place in the Google Landmark Retrieval Competition 2020 [1].*

## 1. Introduction

The Google Landmark Dataset(GLD) V2 is currently the largest public image retrieval and recognition dataset [2], including 4M training data, more than 100,000 query images and nearly 1M index data. But This dataset also contains a lot of noise data. So this year the GLD v2 clean dataset [3] is proposed as the official data of the competition, including 1.5M training data and more than 80000 classes.

The most important change of the competition this year is that participants can only submit tensorflow models while the searching and ranking process are performed by organizers without post-process, local feature and spatial verification. So the main task is training models to extract distinguish global descriptors using large landmark datasets. In our solution, GLDv1 and GLDv2-clean datasets are used and models are trained by PaddlePaddle and Pytorch.

The paper is listed as follows, Section2 will give an overview of our retrieval method and Section3 describes the training, testing and converting strategies in detail. Scores of our models in different stages are also presented.

---

*These authors contributed equally to this work.
[1] https://github.com/PaddlePaddle/PaddleClas
[2] https://github.com/feymanpriv/pymetric

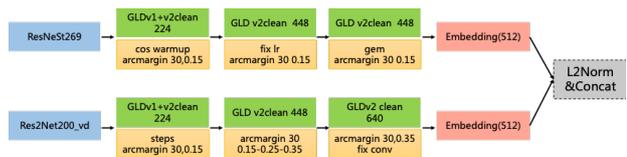

Figure 1. Retrieval method overview

## 2. Retrieval method

Our retrieval method for this competition is depicted in Figure 1. We mainly train two models for final submission and each model includes a backbone model for feature extraction and a head model for classification. ResNeSt269[3] and Res2Net200_vd are selected as the backbone model since their good performance on ImageNet. Head model includes a pooling layer and two fully connected(fc) layers. The first fc layer is often called embedding layer or whitening layer whose output size is 512. While the output size of second fc layer is corresponding to the class number of training dataset. Instead of using softmax loss for training, we train these models with arcmargin loss [4]. Arcmargin loss is firstly employed in face recognition, we found it works well in retrieval tasks which can produce distinguishing and compact descriptor in landmark.

The training process mainly consists of three steps. Firstly, we train the two models with resolution 224x224 on GLDv1 dataset which has total 1215498 images of 14950 classes, and GLDv2-clean dataset which has total 1580470 images of 81313 classes. Secondly, these two models are further trained on GLDv2-clean dataset with resolution 448x448, the parameters of arcmargin loss may change during the process. We believe that using large input size is beneficial to extract feature of tiny landmark. However, we have to adopt the training strategy "from small to large" mainly due to the large cost and lack of GPUs. In the final step, some tricks are experimented to increase the performance. We have tried a lot of methods, such as triplet loss finetuning, circle loss finetuning and etc but only "Gem-Pool" [5] and "Fix" [6] strategy are helpful. Details are described in section3.

---

[3] https://github.com/zhanghang1989/ResNeSt

| Backbone | Init_lr | Batch_size | Learning rate scheduler | Arcmargin-scale | Arcmargin-margin | Resolution |
|---|---|---|---|---|---|---|
| ResNeSt269 | 0.01 | 256 | Cosine-decay | 30 | 0.15 | 224→448 |
| Res2Net200_vd | 0.01 | 256 | Steps decay | 30 | 0.15→0.25→0.35 | 224→448 |

Table 1. Training details of different backbones.

| Backbone | Test resolution 448x448 | | Test resolution 640x640 | | Test resolution 640x640(Fix) | |
|---|---|---|---|---|---|---|
| | Public-score | Private-score | Public-score | Private-score | Public-score | Private-score |
| ResNeSt269 | 0.36972 | 0.33015 | 0.39040 | 0.34718 | —— | —— |
| Res2Net200_vd | 0.36507 | 0.32420 | 0.37115 | 0.33287 | 0.37480 | 0.33376 |

Table 2. Submission mAP@100 scores of different backbones.

For submission, we enlarge the image size to 640, and convert the trained model to tensorflow2.2 format via ONNX. After obtaining the descriptors of embedding layer of above two models, we normalize the descriptors and then concatenate them together. Post-process is finished in tensorflow2.2 models and final descriptor size is 1024.

## 3. Details

### 3.1. Training details

As mentioned above, we trained the network with arcmargin loss. At the data level, we first used GLDv1 and GLDv2-clean data with small resolution to train at a large learning rate, and then used GLDv2-clean data with large resolution to train at a small learning rate. In order to make the final combine result more effective, the details of training the two networks are different. The specific details are listed in Table 1.

During training, we follow practice and perform standard data augmentation with random cropping to 224×224 pixels (448×448 for training with a smaller learning rate) and random horizontal flipping. Input images are normalized through mean channel subtraction. In the experiments, SGD is adopted with weight decay 0.0001, momentum 0.9, and a mini-batch of 256 on 8 V100 GPUs. When the steps learning rate scheduler method is used, the learning rate decay factor is 0.1, and the decay is performed whenever the loss no longer declines, and the margin value is increased at the same time. During training Res2Net200_vd, the resolution was enlarged when the learning rate declined for the first time. During training ResNeSt269, 100 epochs were trained in the first step and then resolution was enlarged to continue training for about 30 epochs.

Table 2 shows the results of training with the above strategies. It can be seen that the two different training strategies have competitive results. Finally we used the "Fix" strategy to continue training the non-conv layer of Res2Net200_vd with resolution 640x640 and the image preprocess method in test period. From the 4th column, we can see the mAP score was further improved.

### 3.2. Testing details

We added a 512-dimensional embedding layer after GAP of both models, and arcmargin loss was connected after this layer. In the infer phase, we performed the L2-Norm operation on the features of the embedding layer and used it as the final feature. The image preprocess in the infer phase is to resize the image to a large size AxA, and then use the center-crop to size BxB, where B/A=0.9201.

### 3.3. Model conversion

Since we not use the tensorflow to train model, we need to convert the models trained by other frameworks to the tensorflow model. Specifically, the ResNeSt269 model trained by Pytorch is converted to the onnx model using onnx[4]. Res2Net200_vd model trained by PaddlePaddle is converted to onnx model using X2paddle[5]. To convert the onnx model to the tf model, we use onnx-tf[6].

| ensemble methods | Resolution | Public score | Private score |
|---|---|---|---|
| ensemble v1(L2-norm+concat) | 448 | 0.38516 | 0.34600 |
| ensemble v2(L2-norm+concat) | 576 | 0.39593 | 0.35376 |
| ensemble v3(L2-norm+concat) | 640 | 0.39992 | 0.36148 |
| **ensemble v4(v3+Res2Net200_vd Fix)** | **640** | **0.40176** | **0.36278** |

Table 3. Submission mAP@100 scores of different ensemble methods.

---

[4] https://github.com/onnx/onnx
[5] https://github.com/PaddlePaddle/X2Paddle
[6] https://github.com/onnx/onnx-tensorflow

### 3.4. Model ensemble

As we all know, model ensemble can stably improve the result. So we finally ensemble the models mentioned above. Specifically, we first perform the L2-Norm operation on the 512-dimensional features of the two models, and then perform the concatenate operation on the two 512-dimensional features in the converted tf2.2 models. Table 3 lists the mAP@100 score of model ensemble.

## 4. Conclusion

In this paper, we present a model training scheme for large-scale images retrieval by team bysj, including training arcmargin loss with adaptive margin values and increasing resolution step by step. And this method allowed us to achieve 2nd place in the Google Landmark Retrieval Competition 2020.